\documentclass[10pt,twocolumn,letterpaper]{article}

\usepackage{cvpr}              
\usepackage{booktabs}
\usepackage{tabularx}
\usepackage{threeparttable}
\usepackage{balance}

\usepackage{fontawesome5}

\usepackage{threeparttable}
\usepackage[utf8]{inputenc}
\usepackage[T1]{fontenc}
\usepackage[utf8]{inputenc}
\usepackage[T1]{fontenc}
\usepackage{newunicodechar}

\newunicodechar{─}{-}   
\newunicodechar{│}{|}   
\newunicodechar{└}{+}   
\newunicodechar{├}{|}
\newunicodechar{₂}{$_2$} 
\usepackage{url}
\usepackage{booktabs}
\usepackage{amsfonts}
\usepackage{nicefrac}
\usepackage{multirow} 
\usepackage{microtype}
\usepackage{xcolor}
\usepackage{graphicx}
\usepackage{amsmath}
\usepackage{enumitem}
\usepackage[hang,flushmargin]{footmisc}

\definecolor{maroon}{cmyk}{0,0.87,0.68,0.32}
\definecolor{myyellow}{RGB}{218, 160, 109}
\definecolor{brickred}{rgb}{0.8, 0.25, 0.33}
\definecolor{brandeisblue}{rgb}{0.0, 0.44, 1.0}
\definecolor{applegreen}{rgb}{0.55, 0.71, 0.0}
\definecolor{aogreen}{rgb}{0.0, 0.5, 0.0}

\definecolor{gdmb}{RGB}{47, 114, 173}  
\definecolor{gdmr}{RGB}{199, 100,  38}
\definecolor{gdmg}{RGB}{70, 155, 118}
\definecolor{gdmm}{RGB}{193, 126, 165}
\definecolor{gdmy}{RGB}{239, 227,  98}
\definecolor{gdmc}{RGB}{110, 179, 228}
\definecolor{gdmk}{RGB}{20, 20, 20}
\definecolor{turquoise}{cmyk}{0.65,0,0.1,0.3}
\definecolor{purple}{rgb}{0.65,0,0.65}
\definecolor{dark_green}{rgb}{0, 0.5, 0}
\definecolor{orange}{rgb}{0.8, 0.6, 0.2}
\definecolor{red}{rgb}{0.8, 0.2, 0.2}
\definecolor{darkred}{rgb}{0.6, 0.1, 0.05}
\definecolor{blueish}{rgb}{0.0, 0.3, .6}
\definecolor{light_gray}{rgb}{0.7, 0.7, .7}
\definecolor{pink}{rgb}{1, 0, 1}
\definecolor{greyblue}{rgb}{0.25, 0.25, 1}
\definecolor{orgred}{rgb}{1.0, 0, 0}

\usepackage{pifont}
\newcommand{\cmark}{\ding{51}}%
\newcommand{\xmark}{\ding{55}}%

\definecolor{sh_gray}{rgb}{0.84,0.84,0.84}
\definecolor{sh_gray2}{rgb}{1,0.89,0.75}
\definecolor{color3}{rgb}{0.95,0.95,0.95}
\definecolor{color4}{rgb}{0.94,0.94,1}
\definecolor{color5}{rgb}{1,0.96,0.88}

\def\dbname{4KLSDB}

\usepackage{cuted}
\usepackage{capt-of}
\definecolor{cvprblue}{rgb}{0.21,0.49,0.74}
\usepackage[pagebackref,breaklinks,colorlinks,allcolors=cvprblue]{hyperref}

\usepackage[capitalize]{cleveref}
\crefname{section}{Sec.}{Secs.}
\Crefname{section}{Section}{Sections}
\Crefname{table}{Table}{Tables}
\crefname{table}{Tab.}{Tabs.}

\def\paperID{*****} 
\def\confName{CVPR}
\def\confYear{2026}

\title{\dbname: A Large-Scale Dataset for 4K Image Restoration and Generation}

\author{
Zihao Zhu$^{1}$ \quad
Kuan-Ru Huang$^{1}$ \quad
Zhaoming Xu$^{1}$ \quad
Renjie Li$^{1}$ \quad
Bo Wu$^{1}$\\
Ruizheng Bai$^{1}$ \quad
Mingyang Wu$^{1}$ \quad
Sayak Paul$^{2}$ \quad
Zhengzhong Tu$^{1*}$\\[0.15cm]
$^{1}$Texas A\&M University \qquad
$^{2}$Hugging Face \\
{\tt\small \{zzh021015, tzz\}@tamu.edu}\\
{\small $^{*}$Corresponding author}
}

\begin{document}
\maketitle
\vspace*{-2.9cm}  

\begingroup
\setlength{\stripsep}{0pt}
\begin{strip}
\centering
\vspace*{-0.3em}

\includegraphics[width=0.99\textwidth]{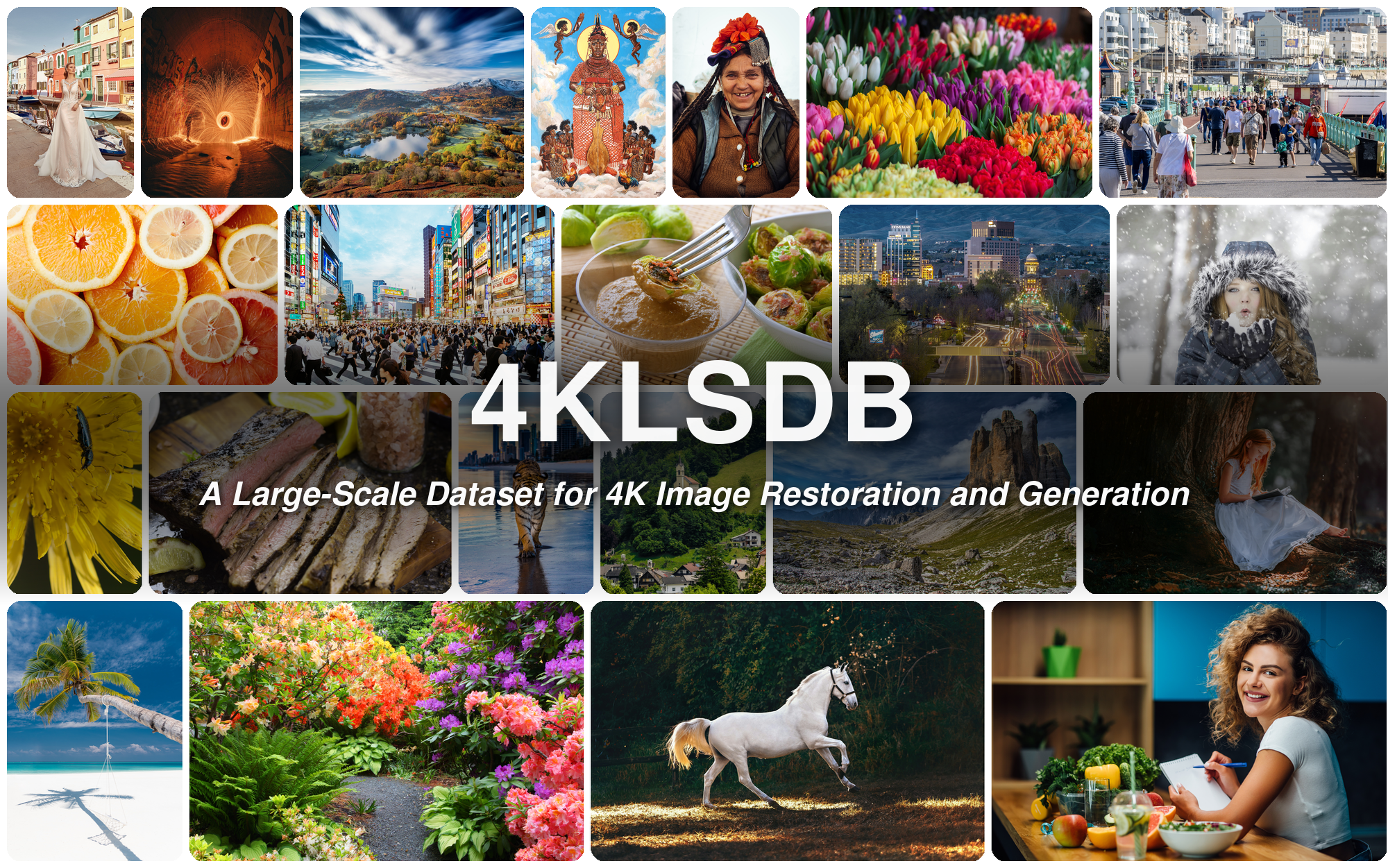}\par

\vspace{0.18em}

\noindent\parbox{0.98\textwidth}{%
\footnotesize
\textbf{Teaser.}
Representative native-4K samples from \dbname{}, covering natural landscapes,
urban scenes, people, food, CGI, artwork, and other high-resolution visual content.
}\par

\vspace{0.65em}
\end{strip}
\endgroup

\begin{abstract}

High-resolution datasets are essential for advancing super-resolution (\emph{SR}) and text-to-image (\emph{T2I}) diffusion research. However, current publicly available datasets lack both the native 4K resolution and the extensive scale necessary for training state-of-the-art models. To address this gap, we introduce a \textbf{4K} \textbf{L}arge \textbf{S}cale \textbf{D}ataset and \textbf{B}enchmark (\emph{4KLSDB}), a large-scale, diverse dataset consisting of 129,484 carefully curated 4K resolution images spanning multiple categories such as nature, urban scenes, people, food, artwork, and CGI, alongside distinct validation and test sets containing 2,000 and 1,984 images respectively. Images were sourced from established open datasets including Photo Concept Bucket, Laion2B, and PD12M. 4KLSDB underwent rigorous multi-stage automated filtering and annotation pipelines involving both human annotators and Large Multimodal Models (\emph{LMMs}) to ensure high aesthetic quality and dataset consistency. We demonstrate 4KLSDB’s effectiveness by training representative super-resolution and diffusion models, observing significant improvements in performance on native 4K benchmarks. Comprehensive experiments illustrate a positive correlation between training on true 4K resolution data and improved fidelity in image restoration task, especially on 4K resolution. We provide the research community a valuable resource to drive progress toward genuinely high-fidelity image synthesis and restoration by providing 4KLSDB. Our project page is available at: \href{https://4klsdb.github.io/}{\texttt{https://4klsdb.github.io/}}.
\end{abstract}


\section{Introduction}






Publicly available large-scale, high-quality native-4K datasets remain scarce, limiting progress in data-driven high-resolution vision. This limitation is particularly pronounced in image restoration, especially super-resolution (SR)~\cite{ledig2017photorealisticsingleimagesuperresolution, li2024systematicsurveydeeplearningbased, wang2018esrganenhancedsuperresolutiongenerative} and related inverse problems such as denoising~\cite{1467423, Zhang_2017, Zhang_2018, ho2020denoisingdiffusionprobabilisticmodels} and deblurring~\cite{5995521,nah2018deepmultiscaleconvolutionalneural, zamir2022restormerefficienttransformerhighresolution}, where higher-resolution and more diverse training data generally lead to sharper reconstructions and stronger generalization~\cite{Li_2023_CVPR,liang2021swinirimagerestorationusing}. A similar challenge also arises in generative modeling, particularly text-to-image diffusion systems, whose ability to synthesize \(2048^2\) or \(4096^2\) images depends critically on access to native high-resolution training examples~\cite{ramesh2022hierarchicaltextconditionalimagegeneration, rombach2022sd}. However, most existing public datasets remain centered on HD or 2K imagery, creating a fundamental bottleneck for both 4K restoration and 4K generation. This bottleneck has also become increasingly visible in recent high-resolution visual systems, including agentic image upscaling and editing~\cite{zuo20254kagentagenticimage4k, ye2026agentbananahighfidelityimage}, interactive video super-resolution~\cite{yu2026sparkvsrinteractivevideosuperresolution}, ultra-high-resolution video generation~\cite{ye2025supergenefficientultrahighresolutionvideo}, and video editing benchmarks~\cite{gao2026vefxbenchholisticbenchmarkgeneric}, where fine-scale details, local consistency, and perceptual artifacts become substantially more important at 4K resolution.

Existing datasets illustrate this gap clearly. DIV2K~\cite{8014884} provides 1,000 images at 2K resolution, but its scale is limited for modern data-hungry models. LSDIR expands the scale to \textit{87\,k} images, yet remains focused on HD and 2K data~\cite{Li_2023_CVPR}. DIV8K~\cite{9021973} includes images at even higher resolutions, up to 8K, but its overall size is still insufficient for current large-scale training needs. On the generative side, datasets such as DiffusionDB~\cite{wang2023diffusiondblargescalepromptgallery} and HQ-Edit~\cite{hui2024hqedithighqualitydatasetinstructionbased} provide image--text pairs, but they rarely exceed \(1024^2\) pixels and do not offer paired low-resolution/high-resolution (LR/HR) data required by SR research. Taken together, existing resources are fragmented: some support restoration but lack native-4K scale, while others support generation but are not designed as public 4K benchmarks spanning both restoration and generation settings. As a result, many recent studies still rely on synthetic upscaling or private collections, which hinders reproducibility and fair comparison.

To address this gap, we introduce \textbf{4KLSDB}, a curated dataset of \textit{129\,k} native-resolution 4K photographs and illustrations, together with 2,000 validation images and 1,984 test images. 4KLSDB covers diverse visual categories, including nature, urban scenes, people, food, artwork, and CGI, as well as multiple shot scales such as long shot, medium shot, close-up, and extreme close-up, annotated using a vision-language model (VLM). To ensure both visual fidelity and data reliability, we build a multi-stage curation pipeline that combines automated heuristics, large multimodal model (LMM) scoring, and human verification to filter out upscaled samples, severe artifacts, and low-quality images. Our main contributions are as follows:
\begin{itemize}[leftmargin=*]
  \item \textbf{4KLSDB}: a large-scale public native-4K image dataset designed to support both image restoration and image generation research, with \textit{129\,k} training images.
  \item A \textbf{robust filtering and quality-control pipeline} that combines rule-based checks, LMM-based aesthetic scoring, and human vetting to remove upscaled or low-quality samples with minimal manual effort.
  \item A \textbf{comprehensive 4K restoration benchmark} with paired LR/HR evaluation sets, supporting both pixel-regression SR models and diffusion-based SR models.
  \item \textbf{Aligned 4K image--text pairs} providing a useful resource for future studies in text-to-image generation, image captioning, and multimodal modeling at ultra-high fidelity.
\end{itemize}

\vspace{0.5em}  
\section{Related Work}

\subsection{Image Restoration Datasets}

Recent progress in image restoration is closely tied to high-quality datasets. Classical SR benchmarks such as DIV2K~\cite{8014884} and LSDIR~\cite{Li_2023_CVPR} have been widely used for training and evaluation, while DIV8K~\cite{9021973} provides higher-resolution images but remains limited in scale. Beyond natural-image SR, SuperBench~\cite{superbench} extends super-resolution evaluation to scientific imaging.

Real-world SR datasets aim to capture degradations from practical imaging pipelines. RealSR-RAW~\cite{peng2024unveiling}, RealSR~\cite{cai2019toward}, and DRealSR~\cite{wei2020component} provide realistic paired LR--HR data, while BSRGAN~\cite{zhang2021designingpracticaldegradationmodel} and Real-ESRGAN~\cite{wang2021realesrgan} study practical degradation synthesis for blind SR. More recently, diffusion-based methods such as SR3~\cite{saharia2021image} and StableSR~\cite{wang2023exploiting} have shown the value of generative priors for perceptual restoration. However, these datasets and methods are still typically trained or evaluated below the native-4K regime, motivating a larger public benchmark tailored to high-resolution restoration and generation.

\begin{figure*}[t]                    
  \centering
  \includegraphics[width=\linewidth]{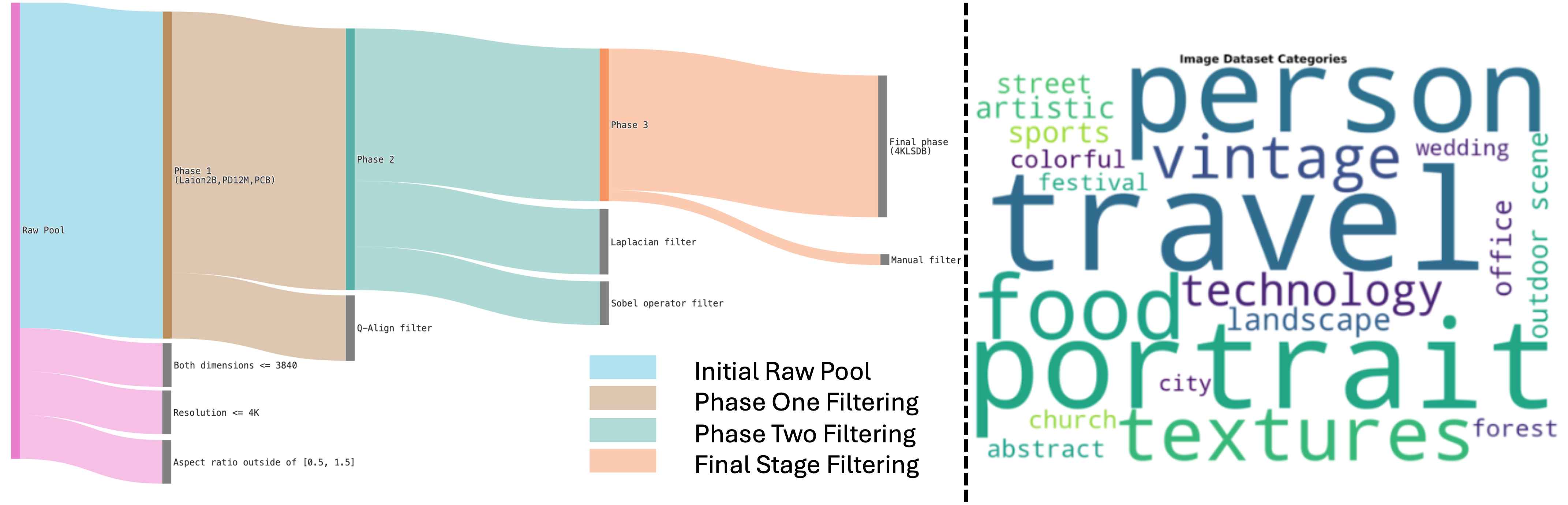}
  \caption{Overview of the 4KLSDB filtering pipeline. An initial raw image pool is progressively refined through automated filters and a final manual inspection stage to obtain a high-quality, aesthetically aligned 4K dataset. The right panel shows the category distribution of the curated data.}
  \label{fig:filtering}
\end{figure*}

\subsection{Text-to-Image Generation Datasets}

Recent work has begun to explore ultra-high-resolution T2I generation and evaluation. Diffusion-4K~\cite{zhang2025diffusion4kultrahighresolutionimagesynthesis} introduces Aesthetic-4K, a curated 4K image--text benchmark for ultra-high-resolution synthesis, while PixArt-$\sigma$~\cite{chen2024pixart} studies efficient 4K generation with a high-resolution evaluation set. Other efforts also move toward high-resolution visual learning: Scaling Vision Pre-Training to 4K Resolution~\cite{shi2025scalingvisionpretraining4k} collects large-scale 1K--4K images for representation learning, Sana~\cite{xie2024sana} demonstrates efficient \(4096\times4096\) generation, and PKU-AIGIQA-4K~\cite{yuan2024pkuaigiqa4kperceptualqualityassessment} provides subjective perceptual-quality labels for 4K AI-generated images.

Despite these advances, existing resources remain fragmented: some target T2I generation only, some focus on evaluation or representation learning, and others rely on partially closed data. They therefore do not provide a unified public native-4K dataset and benchmark for both restoration and generation. Complementary T2I systems and benchmarks, including Imagen~\cite{saharia2022photorealistic}, SDXL~\cite{podell2023sdxl}, JourneyDB~\cite{sun2023journeydb}, GenEval~\cite{ghosh2023geneval}, and T2I-CompBench~\cite{huang2023t2icompbench}, further study photorealistic synthesis, generated-image understanding, and semantic alignment. In contrast, 4KLSDB focuses on native-4K data quality, restoration supervision, and ultra-high-resolution fidelity.

\section{Dataset Description}

We introduce \textbf{4KLSDB}, a large-scale native-4K dataset designed to support both super-resolution (SR) and text-to-image (T2I) generation. In this section, we describe the source datasets, the multi-stage curation pipeline, and the final dataset statistics. Representative examples from 4KLSDB are shown in the teaser at the top of the paper.

\subsection{Source Datasets and Initial Selection}

\paragraph{Source Datasets.}
We begin with several publicly available large-scale image collections and screen them according to their resolution distributions, visual diversity, accessibility, and suitability for downstream curation. Based on this analysis and subsequent manual inspection, we select \emph{LAION-2B}~\cite{laion2B}, \emph{Photo Concept Bucket}, and \emph{PD12M}~\cite{meyer2024pd12m} as the source corpora for 4KLSDB. These datasets provide broad visual coverage and contain sufficient numbers of high-resolution samples while remaining accessible for research use.
\vspace{-3mm}
\paragraph{Resolution-Based Pre-Filtering.}
To construct a candidate pool suitable for native-4K restoration and generation, we first apply the following geometric constraints:
\begin{itemize}[leftmargin=*]
  \item \textbf{Minimum dimension}: at least one image dimension (height or width) must be no smaller than 3840 pixels.
  \item \textbf{Pixel-count requirement}: the total number of pixels must be at least \(3840 \times 2160\).
  \item \textbf{Aspect-ratio constraint}: the aspect ratio must lie within the range \([0.6, 1.6]\) to exclude extreme panoramic or highly elongated images.
\end{itemize}
Only images satisfying all three conditions are retained for the next stage.
\vspace{-3mm}
\paragraph{Automatic Content Annotation.}
After resolution-based filtering, we use Qwen2-VL-7B~\cite{wang2024qwen2vl}\footnote{\url{https://huggingface.co/Qwen/Qwen2-VL-7B}} to annotate each retained image with shot-scale labels (\emph{long shot}, \emph{medium shot}, \emph{close-up}, and \emph{extreme close-up}) and broad content categories (\emph{natural scenes}, \emph{gaming/CGI}, \emph{anime}, and \emph{paintings}). These annotations are used to organize the candidate pool and to monitor content diversity during later split construction. The full source pool and the subset retained after this stage correspond to \emph{Raw Pool} and \emph{Phase 1} in Fig.~\ref{fig:filtering}, respectively.

\subsection{Data Filtering and Processing Pipeline}

After the initial pre-filtering stage, we further refine the candidate pool using perceptual-quality and texture-richness criteria. These stages correspond to Phases 2--3 in Fig.~\ref{fig:filtering}. Our goal is to remove images that technically satisfy the 4K resolution requirement but remain unsuitable for restoration or generation due to poor aesthetics, excessive blur, weak local texture, or severe artifacts.

\subsubsection{Image Quality and Aesthetic Score Filtering}

Resolution alone does not guarantee usable 4K quality. Some images still contain visible compression artifacts, blur, distortion, or poor overall aesthetics despite meeting the pixel-count requirement. To address this issue, we apply an automated scoring stage to approximately \textit{390\,k} Phase-1 images. Specifically, we use Q-Align~\cite{wu2023qalignteachinglmmsvisual} to obtain image quality and aesthetic scores for each sample. We then evaluate multiple retention ratios through visual inspection and select the top 80\% of images as the best trade-off between perceptual quality and dataset size. This stage removes a substantial number of visually unappealing or technically degraded samples before further texture-based filtering.

\subsubsection{Image Richness via Laplacian and Sobel Filtering}

High-resolution SR and 4K T2I generation both benefit from training images with strong local structure, clear edges, and sufficiently rich textures. In contrast, overly flat, blurry, or low-contrast images provide limited supervisory value and can weaken high-frequency learning~\cite{ohtani2024rethinkingimagesuperresolutiontraining}. To suppress such samples, we apply two complementary edge-based filters based on the Laplacian and Sobel operators.

\paragraph{Laplacian Filter.}
We first measure global edge strength using the Laplacian response:
\begin{equation}
L = I * K_{L},
\qquad
K_{L}=
\begin{bmatrix}
0 & 1 & 0\\
1 & -4 & 1\\
0 & 1 & 0
\end{bmatrix},
\end{equation}
where \(I\) denotes the input image and \(L\) is the Laplacian-filtered image. We then compute the variance of the Laplacian response,
\begin{equation}
\operatorname{Var}(L)=
\frac{1}{N}\sum_{x,y}\bigl[L(x,y)-\mu_{L}\bigr]^{2},
\label{eq:lap_var}
\end{equation}
where \(N\) is the total number of pixels and \(\mu_L\) is the mean value of the Laplacian image. Images whose Laplacian variance falls outside an empirically selected interval are removed, since extremely small values typically indicate overly smooth or blurry images, while extreme outliers may correspond to abnormal sharpening or noise.

\paragraph{Sobel-Patch Flatness Ratio.}
To further assess local texture richness, we compute the Sobel gradient magnitude:
\begin{equation}
\begin{aligned}
G_x &= I * K_x,\\
G_y &= I * K_y,\\
M(x,y) &= \sqrt{G_x^2 + G_y^2},
\end{aligned}
\end{equation}
where \(K_x\) and \(K_y\) are the horizontal and vertical Sobel kernels, and \(M(x,y)\) denotes the gradient-magnitude image. We divide \(M\) into non-overlapping \(s \times s\) patches with \(s=240\), and compute the variance of each patch \(P_k\):
\begin{equation}
\operatorname{Var}(P_k)=
\frac{1}{|P_k|}
\sum_{(x,y)\in P_k}\bigl[M(x,y)-\mu_{P_k}\bigr]^{2},
\end{equation}
where \(\mu_{P_k}\) is the mean Sobel magnitude within patch \(P_k\). A patch is considered \emph{flat} if \(\operatorname{Var}(P_k) < T_{\text{flat}}\). We then compute the flat-patch ratio for each image:
\begin{equation}
R_{\text{flat}}=
\frac{1}{N_p}\sum_{k=1}^{N_p}\mathbb{I}\bigl[\operatorname{Var}(P_k)<T_{\text{flat}}\bigr],
\label{eq:sobel_ratio}
\end{equation}
where \(N_p\) is the number of patches. An image is rejected if
\begin{equation}
R_{\text{flat}} \ge T_{\text{ratio}}.
\end{equation}
After pilot filtering experiments, we set \(T_{\text{flat}}=100\) and \(T_{\text{ratio}}=65\%\). Together, the Laplacian and Sobel stages remove images that are excessively flat, blurry, or lacking in local contrast, while preserving visually rich samples for downstream SR and T2I training.

\subsection{Dataset Statistics}

After completing the automated filtering stages, we obtain an interim pool of 134,136 candidate images. To correct residual machine-selection errors, two human annotators review every image using an HTML-based inspection tool and remove 668 samples that are visually unappealing, insufficiently detailed, or otherwise unsuitable. From the remaining verified pool, we construct a validation set of 2,000 images and a test set of 1,984 images, while manually checking category and shot-scale diversity. The remaining 129,484 images form the training set. The final 4KLSDB split therefore contains 129,484 training images, 2,000 validation images, and 1,984 test images, as summarized in Table~\ref{tab:dataset_comparison}.

\begin{table}[t]
  \centering
  \footnotesize
  \setlength{\tabcolsep}{3pt}
  \renewcommand{\arraystretch}{1.05}
  \begin{threeparttable}
  \caption{Comparison of 4KLSDB with existing high-resolution image datasets. ``Native 4K'' indicates that the majority of images have resolutions of at least $3840 \times 2160$ without artificial upscaling.}
  \label{tab:dataset_comparison}

  \begin{tabular*}{\columnwidth}{@{\extracolsep{\fill}}lrrrrc@{}}
    \toprule
    \textbf{Dataset} & \textbf{\#Train} & \textbf{\#Val} & \textbf{\#Test} & \textbf{Max Res.} & \textbf{Native 4K} \\
    \midrule
    DIV2K~\cite{8014884} & 800 & 100 & 100 & 2K & \xmark \\
    LSDIR~\cite{Li_2023_CVPR} & 84,991 & 1,000 & 1,000 & 2K & \xmark \\
    DIV8K~\cite{9021973} & 1,500 & 100 & 100 & 8K & \cmark$^\dagger$ \\
    DiffusionDB~\cite{wang2023diffusiondblargescalepromptgallery} & 14,000,000 & -- & -- & $1024\!\times\!1024$ & \xmark \\
    HQ-Edit~\cite{zhang2025diffusion4kultrahighresolutionimagesynthesis} & $\sim$200,000 & -- & -- & $900\!\times\!900$ & \xmark \\
    \midrule
    \textbf{4KLSDB} & \textbf{129,484} & \textbf{2,000} & \textbf{1,984} & \textbf{4K} & \textbf{\cmark} \\
    \bottomrule
  \end{tabular*}

  \begin{tablenotes}[flushleft]
    \footnotesize
    \item[$\dagger$] DIV8K contains some 8K-resolution images, but its total scale remains relatively limited for large-scale training.
  \end{tablenotes}
  \end{threeparttable}
\end{table}

\section{Benchmark Tasks and Experimental Setup}

We evaluate the proposed 4KLSDB on three tasks: classical super-resolution, real-world blind super-resolution, and 4K text-to-image generation.
To assess the practical value of native-4K supervision, we compare baseline models trained on conventional lower-resolution datasets with the same architectures fine-tuned on 4KLSDB.
For super-resolution tasks, we report both fidelity-oriented and perceptual metrics, while for 4K text-to-image generation we further study whether 4KLSDB improves local detail synthesis and structural coherence in ultra-high-resolution outputs.

\paragraph{Classical Super-Resolution.}
For classical SR, we evaluate three representative restoration models, namely \textbf{HiT-SR}~\cite{aslahishahri2024hitsr}, \textbf{SwinIR}~\cite{liang2021swinirimagerestorationusing}, and \textbf{MambaIR}~\cite{guo2024mambair}, under bicubic downsampling factors of $\{\!\times4,\times8,\times16\}$.
To adapt to GPU memory constraints while preserving native-4K supervision, all models are trained using randomly cropped square patches sampled from the original high-resolution images.

\paragraph{Real-World Super-Resolution.}
For real-world blind SR, we adopt the scale-guided hyper-network blind degradation pipeline~\cite{fu2023scaleguidedhypernetworkblind} and adapt the degradation parameters to the native-4K setting.
We evaluate two representative methods, \textbf{OSEDiff}~\cite{wu2024onestepeffectivediffusionnetwork} and \textbf{SeeSR}~\cite{wu2024seesr}, at upscaling factors of $\{\!\times4,\times8,\times16\}$.
The paired HR/LR test set used in our benchmark is publicly available to support reproducibility.

\paragraph{4K Text-to-Image Generation.}
To further verify the usefulness of 4KLSDB beyond restoration, we fine-tune the text-to-image model \textbf{Sana}~\cite{xie2024sana} on our native-4K image-caption pairs.
This experiment evaluates whether native-4K supervision improve ultra-high-resolution generation quality, especially in terms of local texture fidelity, structural consistency, and visually important fine details.
We compare the original pretrained Sana model with its 4KLSDB fine-tuned counterpart to isolate the effect of native-4K supervision.

\paragraph{Test Splits.}
For classical SR, we report results on both the 4KLSDB and DIV8K~\cite{9021973} test sets to evaluate in-domain performance and cross-dataset generalization.
For real-world SR, we report results on the 4KLSDB test set.
For 4K text-to-image generation, we use a prompt subset selected from the MJHQ-30K benchmark~\cite{li2024playground}, which is also used in Sana's evaluation, and compare the original Sana and the 4KLSDB fine-tuned version under identical inference settings.

\paragraph{Metrics.}
For classical SR, we report PSNR and SSIM~\cite{wang2004image}.
For real-world SR, we report PSNR, SSIM, LPIPS~\cite{zhang2018unreasonable}, DISTS~\cite{ding2020image}, NIQE~\cite{mittal2012making}, and FID~\cite{heusel2017gans} to jointly evaluate distortion fidelity and perceptual quality.
For 4K text-to-image generation, we report both human preference results from a double-blind pairwise user study and patch-based automatic metrics computed on non-overlapping $1024\times1024$ crops, including pCLIPScore~\cite{hessel2022clipscorereferencefreeevaluationmetric} for local text-image alignment and pNIQE~\cite{mittal2012making} for no-reference perceptual quality.

\paragraph{Training Details.}
For fair comparison, each model is evaluated using the same task-specific settings before and after fine-tuning on 4KLSDB.
For 4K text-to-image generation, both Sana variants use identical prompts and the same inference settings.
All experiments are conducted on a system equipped with two NVIDIA A100 GPUs.

\begin{figure*}[!t]
  \centering
  \includegraphics[width=\linewidth]{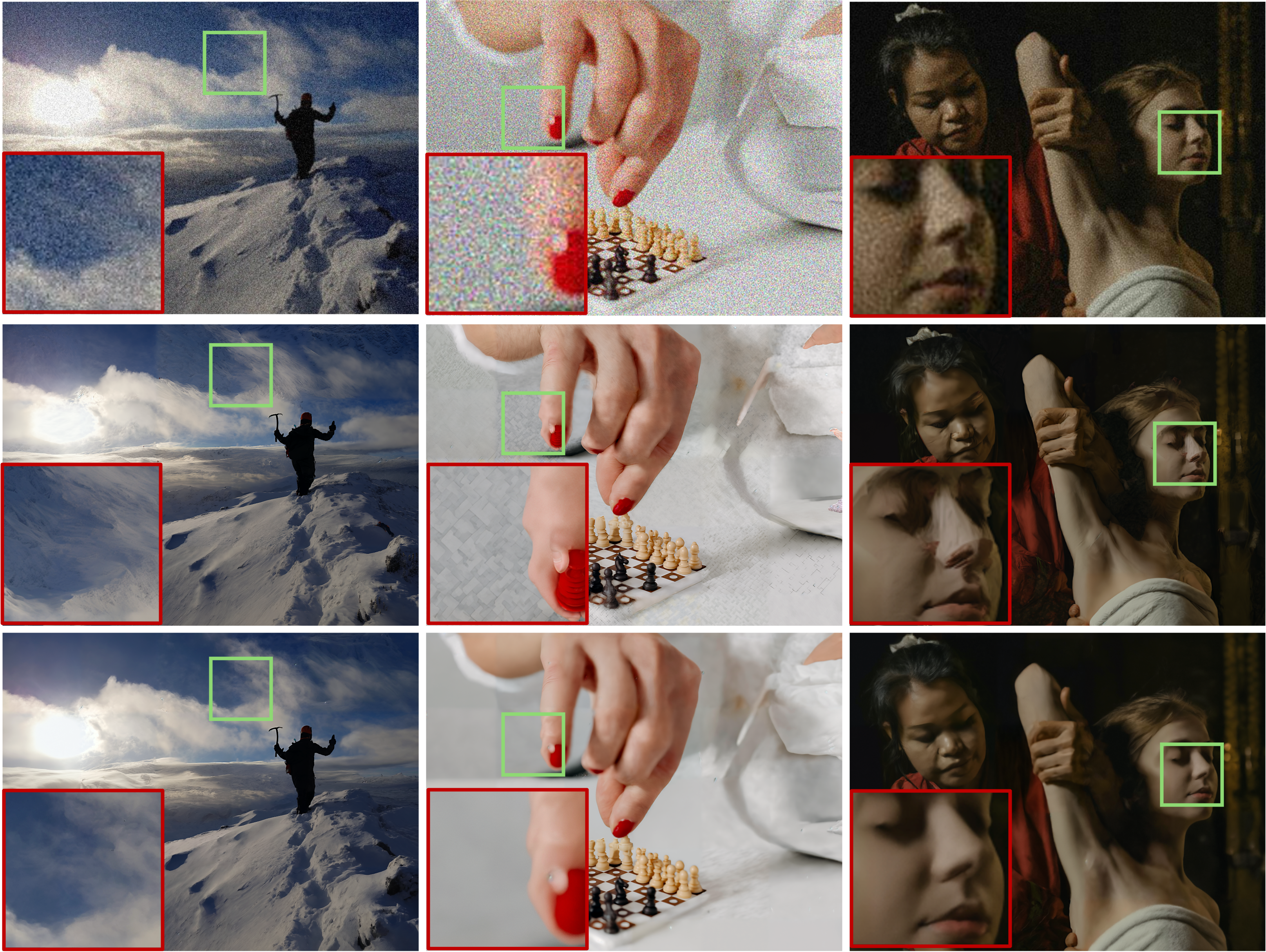}
  \caption{Visual comparison of SeeSR on the 4KLSDB real-SR test set at $\times$4. From top to bottom are the LR input, the original SeeSR baseline, and SeeSR fine-tuned on our 4KLSDB. Fine-tuning with 4KLSDB produces clearer structures and more realistic local details, as highlighted in the red and green inset regions.}
  \label{fig:real_sr_visual}
\end{figure*}

\paragraph{Results.}
Overall, results across classical SR, real-world SR, and 4K text-to-image generation show that fine-tuning on 4KLSDB consistently improves both restoration fidelity and high-resolution visual quality.

\subsection{Classical Super Resolution}

The corresponding results on both the 4KLSDB test set and DIV8K are reported in the following tables.
For HiT-SR, the $\times8$ results are obtained by downsampling the $\times16$ outputs to $\times8$, following the model setting used in our evaluation.

\begin{table}[t]
\centering
\scriptsize
\setlength{\tabcolsep}{4pt}
\renewcommand{\arraystretch}{1.08}
\resizebox{\columnwidth}{!}{%
\begin{tabular}{llcccccc}
\toprule
\textbf{Dataset} & \textbf{Model} 
& \multicolumn{2}{c}{\textbf{$\times$4}} 
& \multicolumn{2}{c}{\textbf{$\times$8}} 
& \multicolumn{2}{c}{\textbf{$\times$16}} \\
\cmidrule(lr){3-4} \cmidrule(lr){5-6} \cmidrule(lr){7-8}
& & \textbf{PSNR}$\uparrow$ & \textbf{SSIM}$\uparrow$
  & \textbf{PSNR}$\uparrow$ & \textbf{SSIM}$\uparrow$
  & \textbf{PSNR}$\uparrow$ & \textbf{SSIM}$\uparrow$ \\
\midrule
4KLSDB & HiT-SR & 24.5027 & 0.6839 & 22.2538 & 0.6394 & 19.4695 & 0.5741 \\
4KLSDB & \textbf{HiT-SR (Ours)} & \textbf{29.2700} & \textbf{0.7896} & \textbf{24.7456} & \textbf{0.6928} & \textbf{23.6919} & \textbf{0.6414} \\
\midrule
DIV8K & HiT-SR & 26.5060 & 0.7987 & 21.9045 & 0.6825 & 19.9915 & 0.6519 \\
DIV8K & \textbf{HiT-SR (Ours)} & \textbf{31.7068} & \textbf{0.8672} & \textbf{23.2167} & \textbf{0.7056} & \textbf{24.6091} & \textbf{0.7107} \\
\bottomrule
\end{tabular}%
}
\caption{Classical HiT-SR evaluation on 4KLSDB and DIV8K. For the $\times$8 setting, the results are obtained by downsampling the $\times$16 outputs to $\times$8.}
\label{tab:hitsr_results}
\end{table}

\begin{table}[t]
\centering
\scriptsize
\setlength{\tabcolsep}{4pt}
\renewcommand{\arraystretch}{1.08}
\resizebox{\columnwidth}{!}{%
\begin{tabular}{llcccccc}
\toprule
\textbf{Dataset} & \textbf{Model} 
& \multicolumn{2}{c}{\textbf{$\times$4}} 
& \multicolumn{2}{c}{\textbf{$\times$8}} 
& \multicolumn{2}{c}{\textbf{$\times$16}} \\
\cmidrule(lr){3-4} \cmidrule(lr){5-6} \cmidrule(lr){7-8}
& & \textbf{PSNR}$\uparrow$ & \textbf{SSIM}$\uparrow$
  & \textbf{PSNR}$\uparrow$ & \textbf{SSIM}$\uparrow$
  & \textbf{PSNR}$\uparrow$ & \textbf{SSIM}$\uparrow$ \\
\midrule
4KLSDB & SwinIR (DIV2K) & 24.11 & 0.6719 & 20.87 & 0.5859 & 19.00 & 0.5656 \\
4KLSDB & SwinIR (DF2K)  & 24.11 & 0.6738 & 20.96 & 0.5915 & 19.20 & 0.5684 \\
4KLSDB & \textbf{SwinIR (Ours)} & \textbf{28.79} & \textbf{0.7774} & \textbf{25.89} & \textbf{0.6877} & \textbf{23.69} & \textbf{0.6376} \\
\midrule
DIV8K & SwinIR (DIV2K) & 24.72 & 0.7610 & 20.89 & 0.6519 & 18.37 & 0.5970 \\
DIV8K & SwinIR (DF2K)  & 24.66 & 0.7632 & 20.46 & 0.6410 & 18.00 & 0.5854 \\
DIV8K & \textbf{SwinIR (Ours)} & \textbf{29.44} & \textbf{0.8314} & \textbf{25.49} & \textbf{0.7260} & \textbf{22.95} & \textbf{0.6640} \\
\bottomrule
\end{tabular}%
}
\caption{Classical SwinIR evaluation on 4KLSDB and DIV8K. Best results in each dataset are shown in bold.}
\label{tab:swinir_results_cvpr}
\end{table}

\begin{table}[t]
\centering
\scriptsize
\setlength{\tabcolsep}{4pt}
\renewcommand{\arraystretch}{1.08}
\resizebox{\columnwidth}{!}{%
\begin{tabular}{llcccccc}
\toprule
\textbf{Dataset} & \textbf{Model} 
& \multicolumn{2}{c}{\textbf{$\times$4}} 
& \multicolumn{2}{c}{\textbf{$\times$8}} 
& \multicolumn{2}{c}{\textbf{$\times$16}} \\
\cmidrule(lr){3-4} \cmidrule(lr){5-6} \cmidrule(lr){7-8}
& & \textbf{PSNR}$\uparrow$ & \textbf{SSIM}$\uparrow$
  & \textbf{PSNR}$\uparrow$ & \textbf{SSIM}$\uparrow$
  & \textbf{PSNR}$\uparrow$ & \textbf{SSIM}$\uparrow$ \\
\midrule
4KLSDB & MambaIR & 25.9182 & 0.7259 & 21.5095 & 0.6382 & 19.4695 & 0.5741 \\
4KLSDB & \textbf{MambaIR (Ours)} & \textbf{30.9249} & \textbf{0.8216} & \textbf{23.8372} & \textbf{0.7195} & \textbf{23.6919} & \textbf{0.6414} \\
\midrule
DIV8K & MambaIR & 26.2929 & 0.7988 & 18.3044 & 0.5613 & 19.7561 & \textbf{0.6507} \\
DIV8K & \textbf{MambaIR (Ours)} & \textbf{31.8728} & \textbf{0.8703} & \textbf{24.0897} & \textbf{0.7121} & \textbf{21.6978} & 0.6078 \\
\bottomrule
\end{tabular}%
}
\caption{Classical MambaIR evaluation on 4KLSDB and DIV8K. Best results in each dataset are shown in bold.}
\label{tab:mambair_results_cvpr}
\end{table}

Across all three architectures, fine-tuning on 4KLSDB consistently improves both PSNR and SSIM on the in-domain 4KLSDB test set and on the cross-dataset DIV8K benchmark.
As shown in table~\ref{tab:hitsr_results}, for HiT-SR the improvement is especially large, with PSNR gains of about $+4.77$\,dB, $+2.47$\,dB, and $+4.22$\,dB on 4KLSDB at $\times4$, $\times8$, and $\times16$, respectively.
A similar trend is observed on DIV8K, showing that the benefit of 4KLSDB is not limited to the training domain.
For SwinIR presented in table~\ref{tab:swinir_results_cvpr}, the 4KLSDB fine-tuned model consistently outperforms both the DIV2K- and DF2K-based baselines across all scales, indicating that native-4K supervision provides substantially stronger high-resolution priors than conventional sub-1K datasets.
MambaIR in table~\ref{tab:mambair_results_cvpr} exhibits the same pattern, with clear gains on both datasets, particularly pronounced at large magnification factors.
Overall, these results demonstrate that training on native 4K data significantly enhances the model's ability to reconstruct fine structures and high-frequency textures, and that these gains become even more pronounced in more challenging $\times8$ and $\times16$ settings.

\begin{figure*}[!t]
  \centering
  \includegraphics[width=\linewidth]{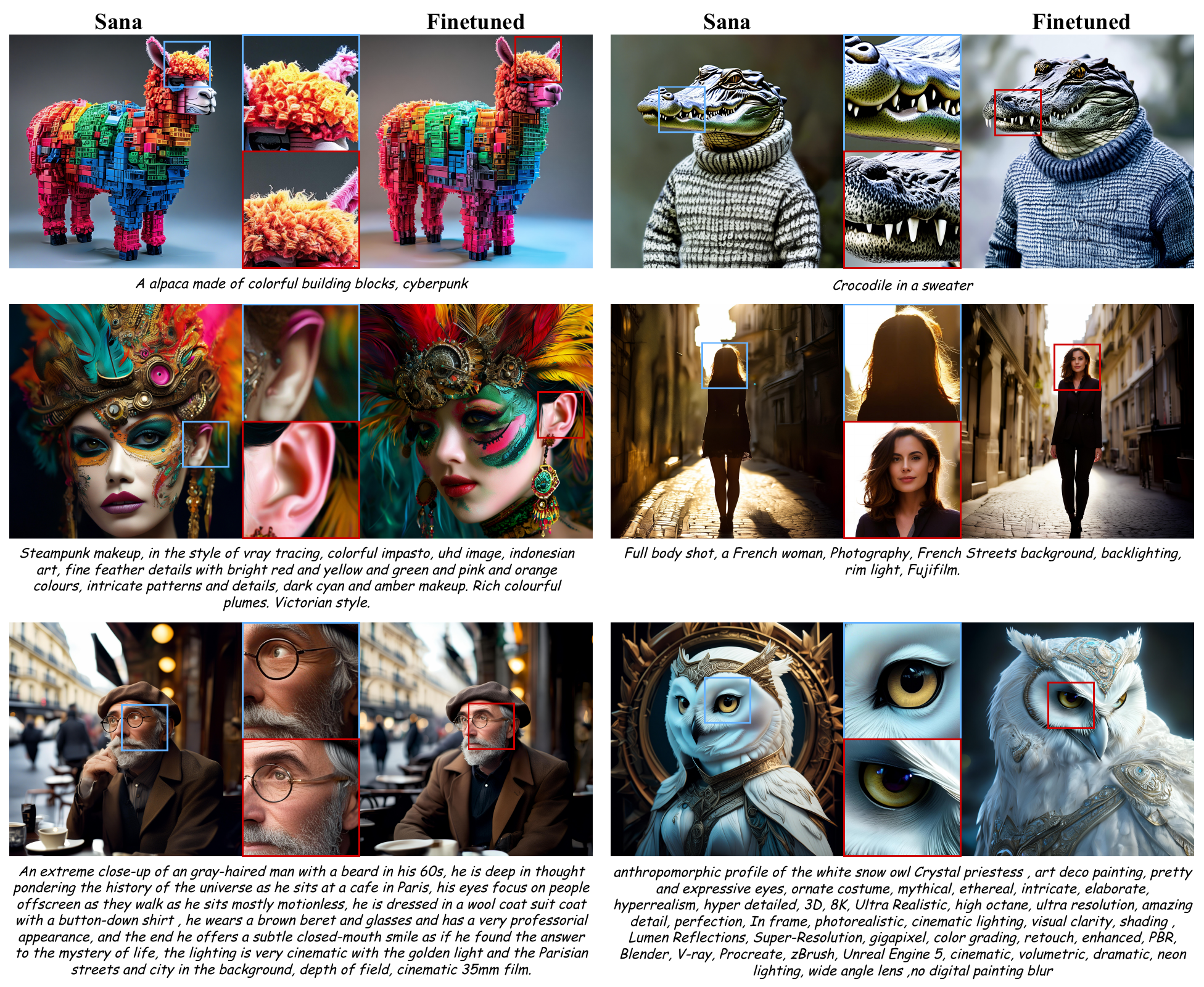}
  \caption{Visual comparison between the original SANA~\cite{xie2024sana} model and the 4KLSDB fine-tuned version under identical prompts.
The fine-tuned model produces clearer local structures, sharper boundaries, and more coherent high-frequency textures in zoomed-in regions.}
  \label{fig:4kt2i}
\end{figure*}

\subsection{Real Super Resolution}

As shown in Table~\ref{tab:real_sr_merged_compact}, similar gains are also observed in the real-world SR setting. We further evaluate real-world blind SR on the 4KLSDB test set using two representative methods, OSEDiff and SeeSR.
Each entry in the table is reported as \textit{Baseline and Ours}, where \textit{Ours} denotes the model fine-tuned on 4KLSDB.
In addition to distortion-oriented metrics, we also report perceptual quality indicators, including LPIPS, DISTS, NIQE, and FID, to provide a more comprehensive evaluation under realistic degradations.

\begin{table*}[t]
\centering
\scriptsize
\setlength{\tabcolsep}{4.0pt}
\renewcommand{\arraystretch}{1.03}
\resizebox{\textwidth}{!}{%
\begin{tabular}{llcccccc}
\toprule
\textbf{Method} & \textbf{Scale}
& \textbf{PSNR}$\uparrow$
& \textbf{SSIM~\cite{1284395}}$\uparrow$
& \textbf{LPIPS~\cite{zhang2018unreasonable}}$\downarrow$
& \textbf{DISTS~\cite{ding2020image}}$\downarrow$
& \textbf{NIQE~\cite{mittal2013niqe}}$\downarrow$
& \textbf{FID~\cite{heusel2017gans}}$\downarrow$ \\
\midrule
OSEDiff & $\times$4
& 27.3589 / \textbf{27.5032}
& 0.7511 / \textbf{0.7568}
& 0.2863 / \textbf{0.2546}
& 0.1604 / \textbf{0.1431}
& 4.3929 / \textbf{4.3142}
& \textbf{28.0671} / 28.3513 \\

OSEDiff & $\times$8
& 23.8554 / \textbf{24.1031}
& 0.6021 / \textbf{0.6188}
& 0.5463 / \textbf{0.4252}
& 0.1833 / \textbf{0.1448}
& \textbf{6.8689} / 7.2678
& 19.5644 / \textbf{17.7353} \\

OSEDiff & $\times$16
& 22.6497 / \textbf{22.6904}
& \textbf{0.6213} / 0.5966
& 0.6571 / \textbf{0.4866}
& 0.2861 / \textbf{0.2170}
& \textbf{6.3227} / 7.3803
& 51.7604 / \textbf{33.9727} \\
\midrule
SeeSR & $\times$4
& 27.0091 / \textbf{28.2485}
& 0.6996 / \textbf{0.7340}
& 0.5231 / \textbf{0.4511}
& 0.1407 / \textbf{0.1272}
& 7.0726 / \textbf{6.1308}
& 38.9548 / \textbf{33.8766} \\

SeeSR & $\times$8
& 24.0986 / \textbf{24.4989}
& 0.6510 / \textbf{0.6713}
& 0.5117 / \textbf{0.4628}
& 0.1607 / \textbf{0.1551}
& 6.8879 / \textbf{6.7440}
& 77.4553 / \textbf{74.4598} \\

SeeSR & $\times$16
& 24.0161 / \textbf{24.4267}
& 0.6810 / \textbf{0.7001}
& 0.5594 / \textbf{0.5197}
& 0.1699 / \textbf{0.1640}
& 7.0108 / \textbf{6.7750}
& 77.4141 / \textbf{74.4033} \\
\bottomrule
\end{tabular}%
}
\caption{Real-SR results on the 4KLSDB test set. Each entry is reported as \emph{Baseline / Ours}, and the better result is highlighted in bold. We keep standard fidelity metrics (PSNR, SSIM, LPIPS, NIQE) and additionally report DISTS and FID, where our method shows stronger and more consistent improvements.}
\label{tab:real_sr_merged_compact}
\end{table*}

Fine-tuning on 4KLSDB also yields clear benefits in the real-SR setting.
For SeeSR, the improvements are highly consistent across all scales and across nearly all reported metrics.
For example, at $\times4$, PSNR improves from $27.0091$ to $28.2485$, SSIM improves from $0.6996$ to $0.7340$, LPIPS decreases from $0.5231$ to $0.4511$, and FID decreases from $38.9548$ to $33.8766$.
The same trend remains visible at $\times8$ and $\times16$, suggesting that 4KLSDB provides useful supervision not only for fidelity restoration but also for perceptual realism under blind degradations.
OSEDiff also benefits from 4KLSDB fine-tuning in most settings, showing improved PSNR, LPIPS, and DISTS across all scales, together with a particularly large FID reduction at $\times16$.
At the same time, some metrics at the most challenging $\times16$ setting remain mixed, such as SSIM and NIQE, which reflects the inherent difficulty of balancing perceptual realism and distortion fidelity in real-world SR.
Qualitative examples in Fig.~\ref{fig:real_sr_visual} further support the quantitative findings, showing that the 4KLSDB fine-tuned model recovers cleaner structures, sharper edges, and more realistic local textures than the original baseline.

\subsection{4K Text-to-Image Generation}
\label{sec:application_4k_t2i}

To verify the usefulness of \textbf{4KLSDB} for generative modeling, we fine-tune \textbf{Sana} on our native-4K image-caption pairs and compare it with the original pretrained model under identical prompts and inference settings. We evaluate both automatic patch-based metrics and a double-blind pairwise user study to assess local detail quality, perceptual realism, and text-image alignment in the 4K regime.

\begin{table}[t]
\centering
\small
\setlength{\tabcolsep}{5pt}
\caption{Quantitative comparison for 4K text-to-image generation.
All images are generated using the same prompts and inference settings.
Patch-based metrics are computed by splitting each 4K output into non-overlapping $1024\times1024$ crops.
Higher is better for pCLIPScore, while lower is better for pNIQE.}
\label{tab:t2i_auto_metrics}
\begin{tabular}{lcc}
\toprule
Model & pCLIPScore $\uparrow$ & pNIQE $\downarrow$ \\
\midrule
Sana (baseline) & 28.62 & 5.21 \\
Sana + 4KLSDB & \textbf{29.27} & \textbf{4.63} \\
\bottomrule
\end{tabular}
\end{table}

\begin{table}[t]
\centering
\small
\setlength{\tabcolsep}{4pt}
\renewcommand{\arraystretch}{1.05}
\caption{Double-blind pairwise user study for 4K text-to-image generation.
We report the preference win rate of the 4KLSDB fine-tuned Sana model over the original Sana baseline.}
\label{tab:t2i_user_study}
\begin{tabular}{ccccc}
\toprule
Overall$\uparrow$ & Detail$\uparrow$ & Real.$\uparrow$ & Artif.$\uparrow$ & Align.$\uparrow$ \\
\midrule
\textbf{57.34\%} & \textbf{60.89\%} & \textbf{74.27\%} & \textbf{64.40\%} & \textbf{52.29\%} \\
\bottomrule
\end{tabular}
\end{table}

As shown in Table~\ref{tab:t2i_auto_metrics}, fine-tuning Sana on \textbf{4KLSDB} consistently improves patch-based automatic metrics over the original baseline. In particular, pCLIPScore increases from 28.62 to 29.27, indicating stronger text-image consistency on local $1024\times1024$ regions, while pNIQE decreases from 5.21 to 4.63, suggesting better perceptual quality and fewer locally visible artifacts. These results show that native-4K supervision benefits not only global generation quality but also the fine-scale visual structures that become critical in ultra-high-resolution outputs.

Table~\ref{tab:t2i_user_study} further confirms this trend from human evaluation. In the double-blind pairwise study, the 4KLSDB fine-tuned model achieves an overall preference rate of 57.34\% over the original Sana baseline. The improvement is especially clear in \textit{detail} (60.89\%), \textit{realism} (74.27\%), and \textit{artifacts} (64.40\%), indicating that raters consistently prefer the fine-tuned model in terms of local sharpness, visual naturalness, and reduced artifact severity. We also observe a smaller but still positive gain in \textit{alignment} (52.29\%), suggesting that the fine-tuned model preserves text-image consistency while mainly improving perceptual quality at high resolution.

Figure~\ref{fig:4kt2i} provides representative qualitative comparisons. Across diverse prompts, fine-tuning on \textbf{4KLSDB} yields sharper boundaries, cleaner structures, and more coherent high-frequency textures in zoomed-in regions. Compared with the original Sana model, the fine-tuned version produces visually stronger local patterns and more stable fine details, which is consistent with both the automatic metrics and the user study results. Overall, these results demonstrate that native-4K supervision from \textbf{4KLSDB} effectively improves ultra-high-resolution text-to-image generation.

\section{Conclusion}

In this paper, we present \textbf{4KLSDB}, a large-scale native-4K image dataset and benchmark designed to support both high-resolution restoration and generative modeling. Unlike conventional sub-1K or 2K resources, 4KLSDB provides native-4K supervision with rich high-frequency details, diverse visual categories, aligned image--caption pairs, and carefully curated validation and test splits.

Extensive experiments on classical super-resolution, real-world blind super-resolution, and 4K text-to-image generation demonstrate the value of 4KLSDB. Across multiple restoration architectures, fine-tuning on 4KLSDB improves reconstruction fidelity and cross-dataset generalization. In real-world SR, it further improves perceptual realism under blind degradations. For 4K text-to-image generation, 4KLSDB improves local detail synthesis, structural consistency, and human preference, showing its usefulness beyond restoration.

Since all validation and test samples are kept at native 4K resolution, 4KLSDB also enables evaluation of scale-dependent artifacts that are often hidden after resizing or low-resolution cropping. This is particularly useful for analyzing over-smoothing, repeated textures, boundary distortions, and other local failures that become more visible under zoomed-in inspection.

Beyond the evaluated tasks, 4KLSDB can also support future high-resolution multimodal research, including detailed captioning, visual question answering, and region-level reasoning, where small objects and fine textures are often lost in lower-resolution data. We hope 4KLSDB will facilitate future research in ultra-high-resolution image restoration, generation, and multimodal understanding.



{\small
\bibliographystyle{ieeenat_fullname}
\bibliography{references}

@inproceedings{Li_2023_CVPR,
  title={Lsdir: A large scale dataset for image restoration},
  author={Li, Yawei and Zhang, Kai and Liang, Jingyun and Cao, Jiezhang and Liu, Ce and Gong, Rui and Zhang, Yulun and Tang, Hao and Liu, Yun and Demandolx, Denis and others},
  booktitle={Proceedings of the IEEE/CVF Conference on Computer Vision and Pattern Recognition},
  pages={1775--1787},
  year={2023}
}

@article{rombach2022sd,
  title   = {High-Resolution Image Synthesis with Latent Diffusion Models},
  author  = {Rombach, Robin and Blattmann, Andreas and Lorenz, Dominik and Esser, Patrick and Ommer, Björn},
  journal = {CVPR},
  year    = {2022}
}

@inproceedings{liang2021swinirimagerestorationusing,
  title={Swinir: Image restoration using swin transformer},
  author={Liang, Jingyun and Cao, Jiezhang and Sun, Guolei and Zhang, Kai and Van Gool, Luc and Timofte, Radu},
  booktitle={Proceedings of the IEEE/CVF international conference on computer vision},
  pages={1833--1844},
  year={2021}
}

@InProceedings{8014884,
	author = {Agustsson, Eirikur and Timofte, Radu},
	title = {NTIRE 2017 Challenge on Single Image Super-Resolution: Dataset and Study},
	booktitle = {The IEEE Conference on Computer Vision and Pattern Recognition (CVPR) Workshops},
	month = {July},
	year = {2017}
}

@article{laion2B,
  title={Laion-5b: An open large-scale dataset for training next generation image-text models},
  author={Schuhmann, Christoph and Beaumont, Romain and Vencu, Richard and Gordon, Cade and Wightman, Ross and Cherti, Mehdi and Coombes, Theo and Katta, Aarush and Mullis, Clayton and Wortsman, Mitchell and others},
  journal={Advances in neural information processing systems},
  volume={35},
  pages={25278--25294},
  year={2022}
}

@article{ramesh2022hierarchicaltextconditionalimagegeneration,
  title={Hierarchical text-conditional image generation with clip latents},
  author={Ramesh, Aditya and Dhariwal, Prafulla and Nichol, Alex and Chu, Casey and Chen, Mark},
  journal={arXiv preprint arXiv:2204.06125},
  volume={1},
  number={2},
  pages={3},
  year={2022}
}

@inproceedings{wang2023diffusiondblargescalepromptgallery,
  title={Diffusiondb: A large-scale prompt gallery dataset for text-to-image generative models},
  author={Wang, Zijie J and Montoya, Evan and Munechika, David and Yang, Haoyang and Hoover, Benjamin and Chau, Duen Horng},
  booktitle={Proceedings of the 61st annual meeting of the association for computational linguistics (volume 1: Long papers)},
  pages={893--911},
  year={2023}
}

@article{hui2024hqedithighqualitydatasetinstructionbased,
  title={Hq-edit: A high-quality dataset for instruction-based image editing},
  author={Hui, Mude and Yang, Siwei and Zhao, Bingchen and Shi, Yichun and Wang, Heng and Wang, Peng and Zhou, Yuyin and Xie, Cihang},
  journal={arXiv preprint arXiv:2404.09990},
  year={2024}
}

@inproceedings{zhang2025diffusion4kultrahighresolutionimagesynthesis,
  title={Diffusion-4k: Ultra-high-resolution image synthesis with latent diffusion models},
  author={Zhang, Jinjin and Huang, Qiuyu and Liu, Junjie and Guo, Xiefan and Huang, Di},
  booktitle={Proceedings of the Computer Vision and Pattern Recognition Conference},
  pages={23464--23473},
  year={2025}
}

@inproceedings{9021973,
  title={Div8k: Diverse 8k resolution image dataset},
  author={Gu, Shuhang and Lugmayr, Andreas and Danelljan, Martin and Fritsche, Manuel and Lamour, Julien and Timofte, Radu},
  booktitle={2019 IEEE/CVF International Conference on Computer Vision Workshop (ICCVW)},
  pages={3512--3516},
  year={2019},
  organization={IEEE}
}

@article{wu2023qalignteachinglmmsvisual,
  title={Q-align: Teaching lmms for visual scoring via discrete text-defined levels},
  author={Wu, Haoning and Zhang, Zicheng and Zhang, Weixia and Chen, Chaofeng and Liao, Liang and Li, Chunyi and Gao, Yixuan and Wang, Annan and Zhang, Erli and Sun, Wenxiu and others},
  journal={arXiv preprint arXiv:2312.17090},
  year={2023}
}

@article{fu2023scaleguidedhypernetworkblind,
  title={Scale guided hypernetwork for blind super-resolution image quality assessment},
  author={Fu, Jun},
  journal={arXiv preprint arXiv:2306.02398},
  year={2023}
}

@inproceedings{ohtani2024rethinkingimagesuperresolutiontraining,
  title={Rethinking image super-resolution from training data perspectives},
  author={Ohtani, Go and Tadokoro, Ryu and Yamada, Ryosuke and Asano, Yuki M and Laina, Iro and Rupprecht, Christian and Inoue, Nakamasa and Yokota, Rio and Kataoka, Hirokatsu and Aoki, Yoshimitsu},
  booktitle={European Conference on Computer Vision},
  pages={19--36},
  year={2024},
  organization={Springer}
}

@article{superbench,
  title={SuperBench: A Super-Resolution Benchmark Dataset for Scientific Machine Learning},
  author={Ren, Pu and Erichson, N. Benjamin and Guo, Junyi and Subramanian, Shashank and San, Omer and Lukic, Zarija and Mahoney, Michael W.},
  journal={Data-centric Machine Learning Research},
  volume={2},
  number={8},
  pages={1--45},
  year={2025}
}

@article{peng2024unveiling,
  title={Unveiling hidden details: A raw data-enhanced paradigm for real-world super-resolution},
  author={Peng, Long and Li, Wenbo and Guo, Jiaming and Di, Xin and Sun, Haoze and Li, Yong and Pei, Renjing and Wang, Yang and Cao, Yang and Zha, Zheng-Jun},
  journal={arXiv preprint arXiv:2411.10798},
  year={2024}
}

@inproceedings{chen2024pixart,
  title={Pixart-$\sigma$: Weak-to-strong training of diffusion transformer for 4k text-to-image generation},
  author={Chen, Junsong and Ge, Chongjian and Xie, Enze and Wu, Yue and Yao, Lewei and Ren, Xiaozhe and Wang, Zhongdao and Luo, Ping and Lu, Huchuan and Li, Zhenguo},
  booktitle={European Conference on Computer Vision},
  pages={74--91},
  year={2024},
  organization={Springer}
}

@inproceedings{yuan2024pkuaigiqa4kperceptualqualityassessment,
  title={Pku-aigiqa-4k: A perceptual quality assessment database for both text-to-image and image-to-image ai-generated images},
  author={Yuan, Jiquan and Li, Jihe and Yang, Fanyi and Cao, Xinyan and Che, Jinming and Lin, Jinlong and Cao, Xixin},
  booktitle={Proceedings of the IEEE/CVF International Conference on Computer Vision},
  pages={3331--3340},
  year={2025}
}

@inproceedings{shi2025scalingvisionpretraining4k,
  title={Scaling vision pre-training to 4k resolution},
  author={Shi, Baifeng and Li, Boyi and Cai, Han and Lu, Yao and Liu, Sifei and Pavone, Marco and Kautz, Jan and Han, Song and Darrell, Trevor and Molchanov, Pavlo and others},
  booktitle={Proceedings of the IEEE/CVF Conference on Computer Vision and Pattern Recognition},
  pages={9631--9640},
  year={2025}
}

@article{wu2024onestepeffectivediffusionnetwork,
  title={One-step effective diffusion network for real-world image super-resolution},
  author={Wu, Rongyuan and Sun, Lingchen and Ma, Zhiyuan and Zhang, Lei},
  journal={Advances in Neural Information Processing Systems},
  volume={37},
  pages={92529--92553},
  year={2024}
}

@article{li2024systematicsurveydeeplearningbased,
  title={A systematic survey of deep learning-based single-image super-resolution},
  author={Li, Juncheng and Pei, Zehua and Li, Wenjie and Gao, Guangwei and Wang, Longguang and Wang, Yingqian and Zeng, Tieyong},
  journal={ACM Computing Surveys},
  volume={56},
  number={10},
  pages={1--40},
  year={2024},
  publisher={ACM New York, NY}
}

@inproceedings{ledig2017photorealisticsingleimagesuperresolution,
  title={Photo-realistic single image super-resolution using a generative adversarial network},
  author={Ledig, Christian and Theis, Lucas and Husz{\'a}r, Ferenc and Caballero, Jose and Cunningham, Andrew and Acosta, Alejandro and Aitken, Andrew and Tejani, Alykhan and Totz, Johannes and Wang, Zehan and others},
  booktitle={Proceedings of the IEEE conference on computer vision and pattern recognition},
  pages={4681--4690},
  year={2017}
}

@inproceedings{wang2018esrganenhancedsuperresolutiongenerative,
  title={Esrgan: Enhanced super-resolution generative adversarial networks},
  author={Wang, Xintao and Yu, Ke and Wu, Shixiang and Gu, Jinjin and Liu, Yihao and Dong, Chao and Qiao, Yu and Change Loy, Chen},
  booktitle={Proceedings of the European conference on computer vision (ECCV) workshops},
  pages={0--0},
  year={2018}
}

@inproceedings{1467423,
  title={A non-local algorithm for image denoising},
  author={Buades, Antoni and Coll, Bartomeu and Morel, J-M},
  booktitle={2005 IEEE computer society conference on computer vision and pattern recognition (CVPR'05)},
  volume={2},
  pages={60--65},
  year={2005},
  organization={Ieee}
}

@article{Zhang_2017,
  title={Beyond a gaussian denoiser: Residual learning of deep cnn for image denoising},
  author={Zhang, Kai and Zuo, Wangmeng and Chen, Yunjin and Meng, Deyu and Zhang, Lei},
  journal={IEEE transactions on image processing},
  volume={26},
  number={7},
  pages={3142--3155},
  year={2017},
  publisher={IEEE}
}

@article{Zhang_2018,
  title={FFDNet: Toward a fast and flexible solution for CNN-based image denoising},
  author={Zhang, Kai and Zuo, Wangmeng and Zhang, Lei},
  journal={IEEE Transactions on Image Processing},
  volume={27},
  number={9},
  pages={4608--4622},
  year={2018},
  publisher={IEEE}
}

@article{ho2020denoisingdiffusionprobabilisticmodels,
  title={Denoising diffusion probabilistic models},
  author={Ho, Jonathan and Jain, Ajay and Abbeel, Pieter},
  journal={Advances in neural information processing systems},
  volume={33},
  pages={6840--6851},
  year={2020}
}

@inproceedings{5995521,
  title={Blind deconvolution using a normalized sparsity measure},
  author={Krishnan, Dilip and Tay, Terence and Fergus, Rob},
  booktitle={CVPR 2011},
  pages={233--240},
  year={2011},
  organization={IEEE}
}

@inproceedings{nah2018deepmultiscaleconvolutionalneural,
  title={Deep multi-scale convolutional neural network for dynamic scene deblurring},
  author={Nah, Seungjun and Hyun Kim, Tae and Mu Lee, Kyoung},
  booktitle={Proceedings of the IEEE conference on computer vision and pattern recognition},
  pages={3883--3891},
  year={2017}
}

@inproceedings{zamir2022restormerefficienttransformerhighresolution,
  title={Restormer: Efficient transformer for high-resolution image restoration},
  author={Zamir, Syed Waqas and Arora, Aditya and Khan, Salman and Hayat, Munawar and Khan, Fahad Shahbaz and Yang, Ming-Hsuan},
  booktitle={Proceedings of the IEEE/CVF conference on computer vision and pattern recognition},
  pages={5728--5739},
  year={2022}
}

@inproceedings{zhang2021designingpracticaldegradationmodel,
  title={Designing a practical degradation model for deep blind image super-resolution},
  author={Zhang, Kai and Liang, Jingyun and Van Gool, Luc and Timofte, Radu},
  booktitle={Proceedings of the IEEE/CVF international conference on computer vision},
  pages={4791--4800},
  year={2021}
}

@inproceedings{hessel2022clipscorereferencefreeevaluationmetric,
  title={Clipscore: A reference-free evaluation metric for image captioning},
  author={Hessel, Jack and Holtzman, Ari and Forbes, Maxwell and Le Bras, Ronan and Choi, Yejin},
  booktitle={Proceedings of the 2021 conference on empirical methods in natural language processing},
  pages={7514--7528},
  year={2021}
}

@article{aslahishahri2024hitsr,
  title={Hitsr: A hierarchical transformer for reference-based super-resolution},
  author={Aslahishahri, Masoomeh and Ubbens, Jordan and Stavness, Ian},
  journal={arXiv preprint arXiv:2408.16959},
  year={2024}
}

@inproceedings{guo2024mambair,
  title={Mambair: A simple baseline for image restoration with state-space model},
  author={Guo, Hang and Li, Jinmin and Dai, Tao and Ouyang, Zhihao and Ren, Xudong and Xia, Shu-Tao},
  booktitle={European conference on computer vision},
  pages={222--241},
  year={2024},
  organization={Springer}
}

@inproceedings{wu2024seesr,
  title={Seesr: Towards semantics-aware real-world image super-resolution},
  author={Wu, Rongyuan and Yang, Tao and Sun, Lingchen and Zhang, Zhengqiang and Li, Shuai and Zhang, Lei},
  booktitle={Proceedings of the IEEE/CVF conference on computer vision and pattern recognition},
  pages={25456--25467},
  year={2024}
}

@article{xie2024sana,
  title={Sana: Efficient high-resolution image synthesis with linear diffusion transformers},
  author={Xie, Enze and Chen, Junsong and Chen, Junyu and Cai, Han and Tang, Haotian and Lin, Yujun and Zhang, Zhekai and Li, Muyang and Zhu, Ligeng and Lu, Yao and others},
  journal={arXiv preprint arXiv:2410.10629},
  year={2024}
}

@article{mittal2013niqe,
  title   = {Making a ``Completely Blind'' Image Quality Analyzer},
  author  = {Mittal, Anish and Soundararajan, Rajiv and Bovik, Alan C.},
  journal = {IEEE Signal Processing Letters},
  volume  = {20},
  number  = {3},
  pages   = {209--212},
  year    = {2013},
  month   = mar,
  doi     = {10.1109/LSP.2012.2227726}
}

@article{heusel2017gans,
  title={Gans trained by a two time-scale update rule converge to a local nash equilibrium},
  author={Heusel, Martin and Ramsauer, Hubert and Unterthiner, Thomas and Nessler, Bernhard and Hochreiter, Sepp},
  journal={Advances in neural information processing systems},
  volume={30},
  year={2017}
}

@article{wang2004image,
  title={Image quality assessment: from error visibility to structural similarity},
  author={Wang, Zhou and Bovik, Alan C and Sheikh, Hamid R and Simoncelli, Eero P},
  journal={IEEE transactions on image processing},
  volume={13},
  number={4},
  pages={600--612},
  year={2004},
  publisher={IEEE}
}

@inproceedings{zhang2018unreasonable,
  title={The unreasonable effectiveness of deep features as a perceptual metric},
  author={Zhang, Richard and Isola, Phillip and Efros, Alexei A and Shechtman, Eli and Wang, Oliver},
  booktitle={Proceedings of the IEEE conference on computer vision and pattern recognition},
  pages={586--595},
  year={2018}
}

@ARTICLE{1284395,
  author={Zhou Wang and Bovik, A.C. and Sheikh, H.R. and Simoncelli, E.P.},
  journal={IEEE Transactions on Image Processing}, 
  title={Image quality assessment: from error visibility to structural similarity}, 
  year={2004},
  volume={13},
  number={4},
  pages={600-612},
  doi={10.1109/TIP.2003.819861}}

@article{mittal2012making,
  title={Making a “completely blind” image quality analyzer},
  author={Mittal, Anish and Soundararajan, Rajiv and Bovik, Alan C},
  journal={IEEE Signal processing letters},
  volume={20},
  number={3},
  pages={209--212},
  year={2012},
  publisher={IEEE}
}

@misc{li2024playground,
      title={Playground v2.5: Three Insights towards Enhancing Aesthetic Quality in Text-to-Image Generation}, 
      author={Daiqing Li and Aleks Kamko and Ehsan Akhgari and Ali Sabet and Linmiao Xu and Suhail Doshi},
      year={2024},
      eprint={2402.17245},
      archivePrefix={arXiv},
      primaryClass={cs.CV}
}

@article{ding2020image,
  title={Image quality assessment: Unifying structure and texture similarity},
  author={Ding, Keyan and Ma, Kede and Wang, Shiqi and Simoncelli, Eero P},
  journal={IEEE transactions on pattern analysis and machine intelligence},
  volume={44},
  number={5},
  pages={2567--2581},
  year={2020},
  publisher={IEEE}
}

@misc{gao2026vefxbenchholisticbenchmarkgeneric,
      title={VEFX-Bench: A Holistic Benchmark for Generic Video Editing and Visual Effects}, 
      author={Xiangbo Gao and Sicong Jiang and Bangya Liu and Xinghao Chen and Minglai Yang and Siyuan Yang and Mingyang Wu and Jiongze Yu and Qi Zheng and Haozhi Wang and Jiayi Zhang and Jie Yang and Zihan Wang and Qing Yin and Zhengzhong Tu},
      year={2026},
      eprint={2604.16272},
      archivePrefix={arXiv},
      primaryClass={cs.CV},
      url={https://arxiv.org/abs/2604.16272}, 
}

@misc{yu2026sparkvsrinteractivevideosuperresolution,
      title={SparkVSR: Interactive Video Super-Resolution via Sparse Keyframe Propagation}, 
      author={Jiongze Yu and Xiangbo Gao and Pooja Verlani and Akshay Gadde and Yilin Wang and Balu Adsumilli and Zhengzhong Tu},
      year={2026},
      eprint={2603.16864},
      archivePrefix={arXiv},
      primaryClass={cs.CV},
      url={https://arxiv.org/abs/2603.16864}, 
}

@misc{zuo20254kagentagenticimage4k,
      title={4KAgent: Agentic Any Image to 4K Super-Resolution}, 
      author={Yushen Zuo and Qi Zheng and Mingyang Wu and Xinrui Jiang and Renjie Li and Jian Wang and Yide Zhang and Gengchen Mai and Lihong V. Wang and James Zou and Xiaoyu Wang and Ming-Hsuan Yang and Zhengzhong Tu},
      year={2025},
      eprint={2507.07105},
      archivePrefix={arXiv},
      primaryClass={cs.CV},
      url={https://arxiv.org/abs/2507.07105}, 
}

@misc{ye2025supergenefficientultrahighresolutionvideo,
      title={SuperGen: An Efficient Ultra-high-resolution Video Generation System with Sketching and Tiling}, 
      author={Fanjiang Ye and Zepeng Zhao and Yi Mu and Jucheng Shen and Renjie Li and Kaijian Wang and Saurabh Agarwal and Myungjin Lee and Triston Cao and Aditya Akella and Arvind Krishnamurthy and T. S. Eugene Ng and Zhengzhong Tu and Yuke Wang},
      year={2025},
      eprint={2508.17756},
      archivePrefix={arXiv},
      primaryClass={cs.LG},
      url={https://arxiv.org/abs/2508.17756}, 
}

@misc{ye2026agentbananahighfidelityimage,
      title={Agent Banana: High-Fidelity Image Editing with Agentic Thinking and Tooling}, 
      author={Ruijie Ye and Jiayi Zhang and Zhuoxin Liu and Zihao Zhu and Siyuan Yang and Li Li and Tianfu Fu and Franck Dernoncourt and Yue Zhao and Jiacheng Zhu and Ryan Rossi and Wenhao Chai and Zhengzhong Tu},
      year={2026},
      eprint={2602.09084},
      archivePrefix={arXiv},
      primaryClass={cs.CV},
      url={https://arxiv.org/abs/2602.09084}, 
}

@misc{meyer2024pd12m,
  title={Public Domain 12M: A Highly Aesthetic Image-Text Dataset with Novel Governance Mechanisms},
  author={Meyer, Jordan and Padgett, Nick and Miller, Cullen and Exline, Laura},
  year={2024},
  eprint={2410.23144},
  archivePrefix={arXiv},
  primaryClass={cs.AI}
}

@misc{wang2024qwen2vl,
  title={Qwen2-VL: Enhancing Vision-Language Model's Perception of the World at Any Resolution},
  author={Wang, Peng and Bai, Shuai and Tan, Sinan and Wang, Shijie and Fan, Zhihao and Bai, Jinze and Chen, Keqin and Liu, Xuejing and Wang, Jialin and Ge, Wenbin and Fan, Yang and Dang, Kai and Du, Mengfei and Ren, Xuancheng and Men, Rui and Liu, Dayiheng and Zhou, Chang and Zhou, Jingren and Lin, Junyang},
  year={2024},
  eprint={2409.12191},
  archivePrefix={arXiv},
  primaryClass={cs.CV}
}

@inproceedings{cai2019toward,
  title={Toward Real-World Single Image Super-Resolution: A New Benchmark and A New Model},
  author={Cai, Jianrui and Zeng, Hui and Yong, Hongwei and Cao, Zisheng and Zhang, Lei},
  booktitle={Proceedings of the IEEE/CVF International Conference on Computer Vision},
  pages={3086--3095},
  year={2019}
}

@inproceedings{wei2020component,
  title={Component Divide-and-Conquer for Real-World Image Super-Resolution},
  author={Wei, Pengxu and Xie, Ziwei and Lu, Hannan and Zhan, Zongyuan and Ye, Qixiang and Zuo, Wangmeng and Lin, Liang},
  booktitle={European Conference on Computer Vision},
  pages={101--117},
  year={2020},
  organization={Springer}
}

@inproceedings{wang2021realesrgan,
  title={Real-ESRGAN: Training Real-World Blind Super-Resolution with Pure Synthetic Data},
  author={Wang, Xintao and Xie, Liangbin and Dong, Chao and Shan, Ying},
  booktitle={Proceedings of the IEEE/CVF International Conference on Computer Vision Workshops},
  pages={1905--1914},
  year={2021}
}

@article{saharia2021image,
  title={Image Super-Resolution via Iterative Refinement},
  author={Saharia, Chitwan and Ho, Jonathan and Chan, William and Salimans, Tim and Fleet, David J. and Norouzi, Mohammad},
  journal={IEEE Transactions on Pattern Analysis and Machine Intelligence},
  volume={45},
  number={4},
  pages={4713--4726},
  year={2023}
}

@article{wang2023exploiting,
  title={Exploiting Diffusion Prior for Real-World Image Super-Resolution},
  author={Wang, Jianyi and Yue, Zongsheng and Zhou, Shangchen and Chan, Kelvin C. K. and Loy, Chen Change},
  journal={International Journal of Computer Vision},
  year={2024}
}

@inproceedings{saharia2022photorealistic,
  title={Photorealistic Text-to-Image Diffusion Models with Deep Language Understanding},
  author={Saharia, Chitwan and Chan, William and Saxena, Saurabh and Li, Lala and Whang, Jay and Denton, Emily and Ghasemipour, Seyed Kamyar Seyed and Ayan, Burcu Karagol and Mahdavi, S. Sara and Lopes, Rapha Gontijo and Salimans, Tim and Ho, Jonathan and Fleet, David J. and Norouzi, Mohammad},
  booktitle={Advances in Neural Information Processing Systems},
  volume={35},
  pages={36479--36494},
  year={2022}
}

@misc{podell2023sdxl,
  title={SDXL: Improving Latent Diffusion Models for High-Resolution Image Synthesis},
  author={Podell, Dustin and English, Zion and Lacey, Kyle and Blattmann, Andreas and Dockhorn, Tim and M{\"u}ller, Jonas and Penna, Joe and Rombach, Robin},
  year={2023},
  eprint={2307.01952},
  archivePrefix={arXiv},
  primaryClass={cs.CV}
}

@misc{sun2023journeydb,
  title={JourneyDB: A Benchmark for Generative Image Understanding},
  author={Sun, Keqiang and Pan, Junting and Ge, Yuying and Li, Hao and Duan, Haodong and Wu, Xiaoshi and Zhang, Renrui and Zhou, Aojun and Qin, Zequan and Wang, Yi and Dai, Jifeng and Qiao, Yu and Li, Hongsheng},
  year={2023},
  eprint={2307.00716},
  archivePrefix={arXiv},
  primaryClass={cs.CV}
}

@inproceedings{ghosh2023geneval,
  title={GenEval: An Object-Focused Framework for Evaluating Text-to-Image Alignment},
  author={Ghosh, Dhruba and Hajishirzi, Hannaneh and Schmidt, Ludwig},
  booktitle={Advances in Neural Information Processing Systems},
  volume={36},
  year={2023}
}

@inproceedings{huang2023t2icompbench,
  title={T2I-CompBench: A Comprehensive Benchmark for Open-World Compositional Text-to-Image Generation},
  author={Huang, Kaiyi and Sun, Kaiyue and Xie, Enze and Li, Zhenguo and Liu, Xihui},
  booktitle={Advances in Neural Information Processing Systems},
  volume={36},
  year={2023}
}
}

\end{document}